%% file: root.tex
\documentclass[letterpaper, 10 pt, conference]{ieeeconf/ieeeconf}  

\IEEEoverridecommandlockouts                              

\usepackage{graphics} 
\usepackage{epsfig} 
\usepackage{mathptmx} 
\usepackage{times} 
\usepackage{amsmath} 
\usepackage{amssymb}  
\usepackage{array} 
\usepackage{booktabs} 
\usepackage{multirow} 
\usepackage{subcaption}
\usepackage{tikz} 

\usepackage{xcolor}

\usepackage{hyperref}
\hypersetup{
    colorlinks=true,
    linkcolor=blue,
    citecolor=blue,
    urlcolor=blue,
    filecolor=blue
}

\title{\LARGE \bf
Dynamics Aware Quadrupedal Locomotion via Intrinsic Dynamics Head
}

\author{Aman Arora and Nalini Ratha%
\thanks{$^{1}$Department of Computer Science,
        University at Buffalo, The State University of New York, Buffalo, USA
        {\tt\small \{amanaror,nratha\}@buffalo.edu}}%
}

\begin{document}

\maketitle
\thispagestyle{empty}

\begin{abstract}
Quadrupedal locomotion plays a critical role in enabling agile, versatile movement across complex terrains. Understanding and estimating the underlying physical dynamics are essential for achieving efficient and stable quadrupedal locomotion. We propose a novel training framework for quadrupedal locomotion that enables the Control Policy to understand and reason about physical dynamics. In simulation, we concurrently train an Intrinsic Dynamics (ID) Head that learns state-to-torque dynamics alongside the Control Policy, and we define a dynamics reward enabled by the ID Head that encourages the Policy toward more predictable dynamical behavior. We also provide a mechanism to tune the learned dynamics in the resulting Policy by controlling the training coefficients of the ID Head. Our simulation experiments show that this mechanism drives convergence to better optima across a wide range of standard quadrupedal locomotion rewards, yielding more efficient and smoother policies. Our real-robot experiments demonstrate zero-shot sim-to-real transfer of these improvements, with significant gains in torque efficiency (16.8\%), action rate (18.6\%), and mechanical power (12.8\%), while improving safe torque occupancy by 6.4\%.

\end{abstract}
\pagestyle{empty}


\section{INTRODUCTION}
Quadrupedal locomotion holds the potential to enable exploration across a wide range of terrains that were previously inaccessible to mechanical robots, largely because of the unstructured nature of these environments. One crucial aspect of legged animals and humans is their intuitive understanding of the physical dynamics of their own bodies, which in turn enables them to exhibit more efficient, stable, and smoother locomotion behavior and allows them to navigate effectively in such unstructured environments.
Recent advances in quadrupedal locomotion have demonstrated remarkable capabilities such as rapid locomotion, multiplicity of walking behavior, and navigation in challenging environments~\cite{lee2020learning, margolis2022rapid, kumar2021rma, ji2022concurrent, hwangbo2019learning, agarwal2022legged, nahrendra2023dreamwaq, margolis2022walktheseways, miki2022learning}. The success of reinforcement learning methods in quadrupedal locomotion, combined with state-of-the-art simulation environments, has been the primary driver behind these advances~\cite{rudin2022learning, makoviychuk2021isaac, todorov2012mujoco}. These learned control methods can be classified into two categories based on their action space: position-based control and torque-based control.

The initial successful applications of RL methods in quadrupedal locomotion employed position-based control, where a learned policy outputs target joint positions that are subsequently converted to torques using low-level controllers such as PID controllers~\cite{agarwal2022legged, hwangbo2019learning, kumar2021rma}. While position-based policies are easier to train as they operate in position space—removing the complexities of actuator dynamics and reducing the action space dimensionality—their lack of awareness of actuator dynamics can lead to overreactions, limited compliance, and inefficient or non-smooth behaviors~\cite{chen2023learning, hogan1985impedance}. When a robot's leg becomes stuck, position-based control can cause motors to generate exceptionally high torques, creating safety hazards. Additionally, policies may converge to inefficient walking behaviors such as bounding gaits instead of more energy-efficient trotting gaits for lower speeds~\cite{fu2021minimizing}.

Seminal work \cite{hwangbo2019learning} attempts to address this gap through accurate modeling of actuator dynamics using a small MLP, which is then used in simulation to convert target positions to torques. While this approach results in a more accurate simulation of real actuators, without a learning signal, there is no incentive for the policy to learn actuator dynamics as it is trained in the position action space. Standard approaches~\cite{rudin2022learning, lee2020learning, fu2021minimizing} typically rely on informed rewards such as action smoothness, high torque penalties, and energy penalties to guide policy behavior, yet each reward targets only one aspect of behavior, leaving the policy without a mechanism to reason about physical dynamics.

\input{combined_torque_prediction_figure}
\begin{figure*}[t]
    \centering
    \includegraphics[width=\textwidth]{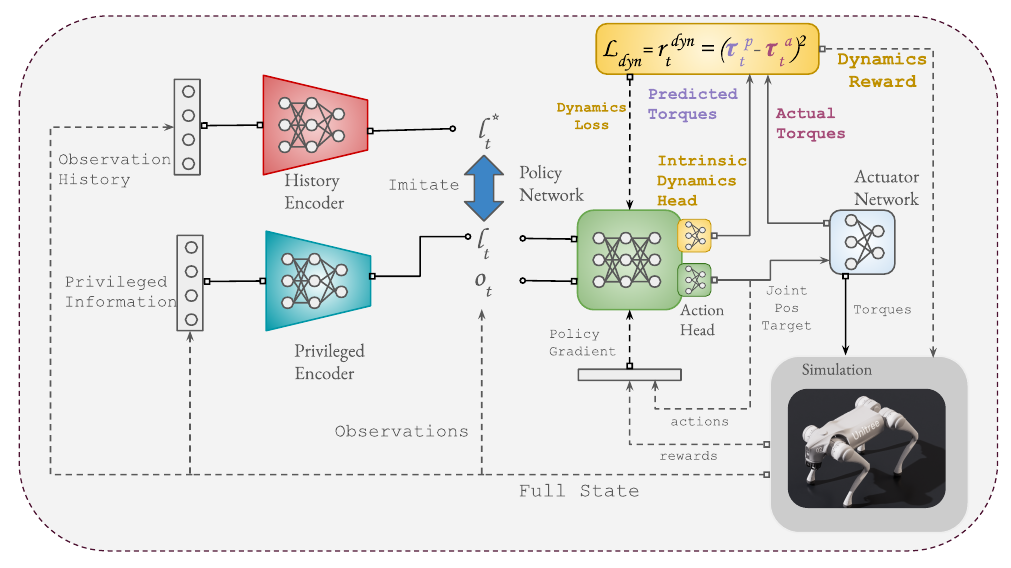}
    \caption{Complete architecture diagram depicting the baseline controller with integrated Intrinsic Dynamics Head. The ID Head processes state information as input and generates torque predictions, which in turn are used to compute the Dynamics Reward and Auxiliary Loss.}
    \label{fig:architecture}
\end{figure*}
In contrast, torque-based learned control methods output torques that are directly applied to actuators. Recent works~\cite{chen2023learning, li2025sata} have demonstrated remarkable progress in torque-based methods, showing high compliance, reduced safety risks, and better generalization to different terrains. Torque-based control is also the preferred choice for model predictive control (MPC)~\cite{DiCarlo2018, chen2023learning}. However, torque-based methods must contend with higher-dimensional action spaces and the nonlinear complexity of state-to-torque dynamics, making them significantly more challenging to train~\cite{kim2023torque}.

Humans perform agile and dynamic skills such as ``holding a cup'' or ``lifting your feet'' in position space, as these tasks are easier to define and reason about in position space. However, we simultaneously possess an intrinsic and intuitive understanding of physical dynamics that makes our motion smoother, more stable, and more efficient. We do not define ``holding a cup'' in terms of force, yet we intuitively know to apply higher or lower forces depending on whether the cup is filled or empty. This observation motivates our hypothesis that an ideal solution should perform and reason about skills in position space while maintaining an intrinsic understanding of torque space and physical dynamics.

Research in self-supervised learning mechanisms~\cite{pathak2017curiosity, agrawal2016poking} has resulted in interesting learning methods that learn world dynamics models in parallel with control policies~\cite{ha2018world,janner2019mbpo,schrittwieser2020mastering}. In curiosity-driven exploration~\cite{pathak2017curiosity, burda2019exploration}, a forward dynamics model predicts the next state given an action and current state, training alongside the policy. An intrinsic reward called curiosity, defined as the prediction error of the forward dynamics model, encourages the policy to explore states where prediction error is high—states that the forward dynamics model predicts incorrectly—resulting in improved exploration.

Inspired by this approach, we introduce an Intrinsic Dynamics Head for quadrupedal locomotion that trains in parallel with the policy, along with an intrinsic dynamics reward that steers the policy toward smoother and more efficient behavior.

The main contributions of this work are:
\begin{itemize}
    \item \textbf{Intrinsic Dynamics (ID) Head:} A state-to-torque dynamics model learned in parallel with the policy, providing a mechanism for reasoning about physical dynamics.
    \item \textbf{Dynamics Reward:} An intrinsic reward that steers the Policy toward smoother and more efficient behavior.
\end{itemize}

We demonstrate significant improvements across a wide category of existing rewards, and we experiment with the real robot showing improvements across multiple efficiency and smoothness metrics. We provide comprehensive training details to ensure reproducibility. 

\section{METHODOLOGY}

In this section, we explain the general method underlying our approach of concurrent policy and Intrinsic Dynamics Head training. First, we detail the baseline architecture of our policy, which employs the student framework~\cite{chen2020learning} for sim-to-real transfer. Then we explain the design of the ID Head, dynamics reward.
Importantly, we maintain architectural similarity with prevalent controllers ~\cite{lee2020learning, kumar2021rma, rudin2022learning,ji2022concurrent} for quadrupedal robots to establish a valid baseline for comparative analysis. This deliberate design choice is evident throughout our baseline controller implementation, from reward structure to neural network architecture, enabling fair and meaningful evaluation of our proposed enhancements.

\subsection{Preliminaries}

We model the environment as a Markov Decision Process (MDP) defined by the tuple $(S, O, A, p, r, \gamma)$: 

Where $\mathbf{a_t} \in A$ is the Action taken at time $\mathbf{t}$; State $\mathbf{s_t} \in S$ is the full state information at time $\mathbf{t}$, where $\mathbf{s_t} = (\mathbf{o_t}, \mathbf{\hat{s}_t})$; $\mathbf{o_t} \in O$ is the observation at time $\mathbf{t}$; $\mathbf{\hat{s}_t}$ is the privileged information at time $\mathbf{t}$; Transition $p : S \times A \rightarrow S$ is the state transition function, where $p(\mathbf{s_{t+1}} | \mathbf{s_t}, \mathbf{a_t})$ is the probability of transitioning from $\mathbf{s_t}$ to $\mathbf{s_{t+1}}$ given action $\mathbf{a_t}$; Reward $r : S \times A \rightarrow \mathbb{R}$, $\mathbf{r_t} = r(\mathbf{s_t}, \mathbf{a_t})$ is the reward received at time $\mathbf{t}$ for taking action $\mathbf{a_t}$ in state $\mathbf{s_t}$; and $\gamma$ is the discount factor.

\subsection{Baseline Policy Architecture}
\subsubsection{\textbf{Observation and State Space}}

The full state at time $\mathbf{t}$ is given by $\mathbf{s_t} = [\mathbf{o_t}, \mathbf{\hat{s}_t}]$, where $\mathbf{o_t}$ is the observation and $\mathbf{\hat{s}_t}$ is the privileged information vector. The privileged information $\mathbf{\hat{s}_t}$ consists of features available only in simulation (e.g., ground-truth dynamics or environment parameters) and is not accessible on the real robot during deployment. This privileged vector is used during training to enable a teacher policy in a privileged learning setup. Table~\ref{tab:privileged_obs} provides the details of the privileged observations used in our approach.

\input{privileged_obs_table}
The observation $\mathbf{o_t}$ at time $\mathbf{t}$ includes the joint positions $\mathbf{q_t}$, joint velocities $\mathbf{\dot{q}_t}$, and the gravity vector projection $\mathbf{g_t}$, which represents the projection of the gravity vector in the robot's frame and serves as an orientation cue. It also contains the previous action $\mathbf{a_{t-1}}$, which is included to improve training stability, as well as the commanded linear velocity $\mathbf{v_t^{\text{cmd}}}$ and commanded angular velocity $\mathbf{\omega_t^{\text{cmd}}}$.

Thus, the observation at time $\mathbf{t}$ is given by:
\[
\mathbf{o_t} = [\mathbf{q_t},  \mathbf{\dot{q}_t},  \mathbf{g_t},  \mathbf{a_{t-1}},  \mathbf{v_t^{\text{cmd}}},  \mathbf{\omega_t^{\text{cmd}}}]
\]

\subsubsection{\textbf{Action Space}}
The policy outputs actions $\mathbf{a_t} \in \mathbb{R}^{12}$, where each element corresponds to a target joint position for one of the robot's 12 actuated joints. We employ an actuator network~\cite{hwangbo2019learning} that maps these position targets to joint torques, thereby modeling the actuator dynamics with greater fidelity, which enables us to further train ID Head on a more reliable torque data. 

\subsubsection{\textbf{Reward Function Design}}

Our reward function closely follows the approach presented in \cite{rudin2022learning, lee2020learning, ji2022concurrent, kumar2021rma} and incorporates insights from \cite{fu2021minimizing}. The reward structure prioritizes linear and angular velocity tracking as primary objectives, complemented by auxiliary rewards for stability and smoothness. Notably, these auxiliary rewards are designed to guide the policy towards desirable behavior such as smoother actions, lower DoF velocities and accelerations, which are most positively affected by our ID Head and dynamics reward. The total reward at each timestep is computed as a weighted sum of individual reward components:

$$
r_{\text{total}} = \sum_{i} w_i \cdot r_i
$$

where $w_i$ are the reward weights and $r_i$ are the individual reward components. Table~\ref{tab:reward_new} provides a comprehensive overview of all reward functions used in our system, with their mathematical expressions and corresponding coefficients.

\input{rewards_active}

\subsubsection{\textbf{Policy Optimization}}
We use a teacher-student training framework \cite{lee2020learning, kumar2021rma, chen2020learning} where the teacher policy is trained in simulation using privileged information and the student learns from the trained teacher policy using imitation learning \cite{ross2011reduction}.

We use a Privileged Encoder 
$$\mu_{\theta_{enc}}(\hat{\mathbf{s}}_t) = \mathbf{l}_t$$ 
similar to \cite{kumar2021rma} to encode privileged information $\hat{\mathbf{s}}_t$ to latent representation $\mathbf{l}_t$. The input to the policy network is then formed as 
$$\mathbf{x}_t = [\mathbf{o}_t, \mathbf{l}_t],$$ 
which is the concatenation of observations and latent encoding. The policy 
$$\pi_{\theta_p}(\mathbf{x}_t) = \mathbf{a}_t$$ 
maps this combined input to actions.

The baseline policy is a multi-layer perceptron with architecture details provided in Table~\ref{tab:architecture}. Our objective is to learn a policy $\pi_{\theta_p}$ and privileged encoder $\mu_{\theta_a}$ that maximize the expected discounted reward starting from state $\mathbf{s_0}$:

$$
J(\theta_p, \theta_{enc}) = \mathbb{E}_{\mathbf{s_0} \sim p(\mathbf{s_0}), \mathbf{a_t} \sim \pi_{\theta_p}([\mathbf{o_t}, \mu_{\theta_{enc}}(\hat{\mathbf{s}}_t)])} \left[ \sum_{t=0}^{\infty} \gamma^t r(\mathbf{s_t}, \mathbf{a_t}) \right]
$$

We train both the policy and privileged encoder using the Proximal Policy Optimization (PPO) algorithm \cite{schulman2017proximal}.
To deploy our policy on real hardware without access to privileged information, we employ a student-teacher framework \cite{chen2020learning}. While the teacher policy uses privileged information during training, the student policy must operate solely on observations available on the real robot. Instead of using a privileged encoder that encodes privileged information, the student policy employs a history encoder that approximates the latent representation $\mathbf{l}_t$ from a history of observations.

\input{architecture}
The History Encoder $\eta_{\theta_h}$ is implemented as a Temporal Convolution Network (TCN)~\cite{bai2018empirical} similar to \cite{lee2020learning} and takes a sequence of past observations $\mathbf{h}_t = [\mathbf{o}_{t-T+1}, \mathbf{o}_{t-T+2}, \ldots, \mathbf{o}_{t-1}, \mathbf{o}_t]$ as input, where $T=100$ is the sequence length.:

$$
\hat{\mathbf{l}}_t = \eta_{\theta_h}(\mathbf{h}_t)
$$

This Head is trained to minimize the following loss function:

$$
\mathcal{L}_{\text{his\_enc}} = \|\mathbf{a}_t - \hat{\mathbf{a}}_t\|^2 + \|\mathbf{l}_t - \hat{\mathbf{l}}_t\|^2
$$

where $\mathbf{a}_t = \pi_{\theta_p}([\mathbf{o}_t, \mu_{\theta_{enc}}(\hat{\mathbf{s}}_t)])$ are the actions produced by the teacher policy and $\hat{\mathbf{a}}_t = \pi_{\theta_p}([\mathbf{o}_t, \hat{\mathbf{l}}_t])$ are the actions produced by the student policy. During this imitation learning phase, the policy parameters $\theta_p$ remain frozen, and only the history encoder parameters $\theta_h$ are updated.

We use the Dataset Aggregation (DAgger) algorithm \cite{ross2011reduction} in an online fashion on trajectories generated by rolling out the student policy. The architecture details for both the policy network and privileged/history encoder modules are provided in Table~\ref{tab:architecture}.

\input{domain_rand}
We also perform domain randomization\cite{peng2018sim} for better sim-to-real transfer, similar to \cite{ji2022concurrent}. The domain randomization parameters, critical for sim-to-real transfer, are detailed in Table~\ref{tab:domain_rand}. This randomization encompasses terrain properties, robot dynamics, and sensor characteristics to ensure robust policy performance across varying conditions.

\subsection{Intrinsic Dynamics Head and Dynamics Reward}
In this section, we describe in detail the Intrinsic Dynamics Head and the dynamics reward. Figure~\ref{fig:architecture} illustrates the overall architecture of our approach.
\subsubsection{\textbf{Intrinsic Dynamics Head}}

We define Intrinsic Dynamics as the mapping from the robot's state $\mathbf{s}_t$ to the torques $\boldsymbol{\tau}_t$ applied by the robot at time $t$. Formally:

$$f(\mathbf{s}_t) = \boldsymbol{\tau}_t$$

This Intrinsic Dynamics Function represents the underlying physical relationship between the robot's state and the resulting torques that are generated during locomotion. We can learn this function in simulation by parameterizing $f$ with $\theta_d$ to minimize the following loss function:
$$
\mathcal{L}_{\text{dyn}} = \alpha_{\text{dyn}} \cdot \|\tau_t^a - \tau_t^p\|^2
$$

Here, $\alpha_{\text{dyn}}$ is a scaling coefficient, $\tau_t^p$ represents the torque values predicted by the parameterized model of $f$, and $\tau_t^a$ corresponds to the actual torques applied during the simulation step. Since we employ actuator networks in simulation to convert actions to torques, the $\tau_t^a$ values closely approximate the torques that would be produced by the real robot for corresponding actions, enhancing the fidelity of our dynamics modeling.
We model the parameters of the Intrinsic Dynamics Function as an additional head on the policy network:  In this architecture, the parameters $\theta_d$ and $\theta_p$ share common representations except for those in their respective output heads. This design is illustrated in Figure~\ref{fig:architecture}, where the policy network simultaneously outputs actions, which are joint position targets, as well as torque predictions.

$$\pi_{\theta_p,\theta_d}(\mathbf{o}_t, \mu_{\theta_{enc}}(\hat{\mathbf{s}}_t)) = [a_t, \tau_t^p]$$

To train the Intrinsic Dynamics Head effectively with the Policy, we incorporate $\mathcal{L}_{\text{dyn}}$ as an auxiliary term in the PPO loss function \cite{schulman2017proximal} used to train the overall policy.
It is important to note that although the policy network has a ID Head, it is used only during training, and during deployment, we simply ignore the torque output or don't deploy the ID Head at all.
\subsubsection{\textbf{Dynamics Reward}}

We define dynamics reward as a penalty on the generation of unpredictable torques. Specifically, given a state $\mathbf{s}_t$ at time $\mathbf{t}$, we penalize how difficult it is to predict the torques generated from actions $\mathbf{a}_t$ outputted by the Policy for this state. Formally:

$$
r_t^{\text{dyn}} = w_{\text{dyn}} \cdot \|\tau_t^a - \tau_t^p\|^2
$$
where $w_{\text{dyn}}$ controls the influence of the dynamics reward on the overall policy optimization process, $\tau_t^p$ represents the predicted torque values, and $\tau_t^a$ corresponds to the actual torques applied during execution. As $r_t^{\text{dyn}}$ is a penalty:
$$w_{\text{dyn}} < 0.0$$
In our experiments, we explore different values for the scaling coefficients $\alpha_{\text{dyn}}$ and $w_{\text{dyn}}$ across various experimental setups. The specific parameter configurations are given in the next section, which explains the effect of the dynamics reward and ID Head on the Policy Training and converged behavior. Their effects on performance are detailed in the Results section.

\input{dynamics_plots}

\subsubsection{\textbf{On-Policy State Distribution Shift}}

As demonstrated in the Results section, the dynamics reward induces a state distribution shift toward smoother and efficient policy regimes, manifesting in improved performance across a wider category of rewards, including stability and smoothness rewards in the reward table. This occurs because, to minimize dynamics reward, the policy converges toward actions that render torques more predictable, consequently yielding well-behaved policies as evidenced by enhanced energy efficiency reflected in power consumption rewards, more optimum action smoothness rewards, and other significant effects detailed in the results section.

This mechanism bears much similarity to curiosity-driven exploration \cite{pathak2017curiosity} which encourages the policy to explore unknown states by rewarding the policy to visit states where the next state prediction error is high for Forward Dynamics Model, but in contrast, our approach penalizes the policy when torque prediction error is high, as we require torques to be predictable for stable and smooth policy behavior. We also penalize the ID Head through $\mathcal{L}_{\text{dyn}}$, which learns to predict more accurately. Consequently, unlike Curiosity-Driven exploration, both the Dynamics Loss and Dynamics Reward signals work towards the same objective. This alignment should be considered carefully as accurate torque prediction is not our primary goal; rather, we aim to guide the policy toward more predictable torque regimes.

However, with this configuration, the ID Head can rapidly learn to predict torques more accurately or be biased towards more non-linear Dynamics of initial training iterations, reducing the reward incentive for the policy to generate actions that render torques predictable. Therefore, a careful tuning of $\alpha_{\text{dyn}}$ and $w_{\text{dyn}}$ described below is necessary. To achieve this, we employ the following three strategies:

\begin{itemize}
    \item \textbf{Adjust the loss coefficient}: Reduce the loss coefficient $\alpha_{\text{dyn}}$ of the Dynamics Loss $\mathcal{L}_{\text{dyn}}$. Reducing the coefficient delays the ID head learning and keeps pressure on the policy to select actions that result in more predictable torques.
    \item \textbf{Adjust the reward weight}: Dynamics Reward weight $w_{\text{dyn}}$ should be adjusted based on the Dynamics Loss ranges produced by the ID Head. If the proportion of the Dynamics Reward is too less in total reward then it won't produce any meaningful effect. On the other hand if the proportion is high then it can worsen the main task performance, hence the values that strike a balance between these two effects are preferred.
    \item \textbf{Zero the loss coefficient}: Another useful way is to completely zero out the loss coefficient $\alpha_{\text{dyn}}$ and tune for $w_{\text{dyn}}$. Setting the coefficient to zero puts the highest pressure on the policy as reward becomes the only signal, forcing the policy to fully rely on action selection to minimize Dynamics Reward penalty $r_t^{\text{dyn}}$.
\end{itemize}
In our experiments we set $\alpha_{\text{dyn}}$ to 3e-4 and $w_{\text{dyn}}$ is set to -1e-2. If Dynamics Loss coefficient $\alpha_{\text{dyn}}$ is set to zero $w_{\text{dyn}}$ is set to -5e-4. These values are sensitive to dynamics randomization settings and are applicable for our Domain Randomization settings described in Table~\ref{tab:domain_rand}. For different Dynamics Randomization settings optimal values for $\alpha_{\text{dyn}}$ and $w_{\text{dyn}}$ can be found by strategies described above.
\input{prediction_head_rewards_table}

\section{EXPERIMENTAL SETUP}

\subsection{Simulation Environment}

We utilize the NVIDIA Isaac Gym SImulation Environment \cite{makoviychuk2021isaac}, a highly parallelized physics simulation framework that enables simultaneous training of 4096 domain-randomized Unitree GO2 quadrupedal robot agents. The terrains consisted of
smooth and rough terrains with 10 levels of progressive hardness. Our implementation follows the approach described in \cite{rudin2022learning}, leveraging the open-source code base for efficient reinforcement learning in legged locomotion tasks.

Training was conducted on an NVIDIA A100 GPU server with 80GB VRAM, although the actual memory requirements were approximately 10-12GB. The training process required 5000 iterations, completing in approximately 5 hours while collecting data equivalent to 160 days of real-world experience.
\input{real_robot_comparison}

\subsection{Real Robot Experimental Setup}

To validate our method on real hardware, we deployed both the baseline controller and the dynamics reward policies on a Unitree GO2 robot. We ran each controller for 2 runs with 10,000 policy steps per run to collect data. Both policies operated at 50 Hz. We collected proprioceptive data, including joint positions, velocities, commands, actions, and torque estimates from the sensors of the Robot through Unitree SDK.

\section{Results}

\subsection{Simulation Results}
Our experimental analysis of training with an integrated Intrinsic Dynamics Head with and without auxiliary loss revealed significant improvements across multiple locomotion metrics when the dynamics reward is used. Our experiments demonstrates substantial performance gains in energy efficiency ($r_{P}$), stability indicator ($r_{v_z}$, $r_{\omega_{xy}}$), and smoothness rewards ($r_{a}$, $r_{\dot{q}}$, $r_{\ddot{q}}$), while maintaining comparable performance on primary task objectives of linear velocity ($r_{v}$) and angular velocity ($r_{\omega}$). 

Figure~\ref{fig:combined_results} illustrates the comparative performance between our approach and the baseline, highlighting the marked improvements in vertical stability ($r_{v_z}$), energy consumption ($r_{P}$), and action smoothness ($r_{a}$) without compromising primary locomotion objectives. The improvements in linear velocity Z are most prominent , which we attribute to the policy optimizing away from jumping-like motions that generate high torque spikes and are inherently less predictable. Table~\ref{tab:prediction_head_rewards} presents a comprehensive analysis of performance metrics with and without the dynamics reward integration. The consistent improvement across diverse reward categories provides compelling evidence for a fundamental state distribution shift induced by the dynamics reward mechanism, directing the policy toward inherently more stable and efficient motion patterns. All these improvements can be understood through the same reasoning: the policy evolves toward behaviors that generate more predictable torque patterns for the ID Head.

\input{real_robot_metrics}

\subsection{Real Robot Results}
Hardware experiments on the Unitree GO2 quadruped demonstrate significant performance improvements with our dynamics-aware approach. The Dynamics Reward Policy exhibits a trotting gait with stable locomotion patterns Figure~\ref{fig:real_robot_comparison} (a). The dynamics reward policy exhibits notable advantages in three key areas: reduced action rates, lower torque magnitudes, and decreased mechanical power consumption. Figure~\ref{fig:real_robot_comparison} (b)-(e) presents a comparative analysis between standard and dynamics reward policies across 250 locomotion steps under identical velocity commands. The data reveal smoother motion profiles, evidenced by reduced joint position error fluctuations, while maintaining comparative locomotion performance with substantially lower energy requirements. We observe that in simulation when velocity commands are changed suddenly, depicted in Figure~\ref{fig:real_robot_comparison} (b), which shows that the commanded x velocity is changed from 0.6 to -0.6, the Baseline Policy changes swiftly in response then readjusts but the transition for the Dynamics Reward Policy is smoother and homogeneous.

Table~\ref{tab:real_robot_metrics} provides a quantitative performance comparison across various evaluation metrics. Particularly noteworthy is the Safe Occupation Zone metric, which measures the percentage of execution steps where torque and torque rate values remain within designated safety thresholds. The dynamics reward policy achieves approximately 6\% improvement in this critical metric, providing empirical validation of our theoretical framework. This improvement confirms the hypothesized on-policy state distribution shift toward more stable locomotion regimes, demonstrating that dynamics-aware policies naturally converge toward smoother and safer motion patterns.

\section{CONCLUSIONS}

In this paper, we presented a novel mechanism for learning state-to-torque dynamics in simulation through an Intrinsic Dynamics (ID) Head and leveraging this dynamics model to guide policy learning toward smoother and more efficient behaviors via a Dynamics Reward. A key advantage of our approach is that it requires no additional computational overhead during deployment, as the dynamics learning mechanism is applied exclusively during simulation while still producing measurable improvements on the real robot.

One limitation of our approach is its reliance on accurate actuator networks~\cite{hwangbo2019learning} that precisely model the physical actuators to generate reliable torque data for ID Head training. However, this requirement is also a contributing factor to the successful transfer of simulation improvements to the real robot. An interesting future direction would be to train the ID Head using torque data from a PD controller and investigate its effects on our mechanism in both simulation and real-world scenarios.
Another limitation is the sensitivity of the Dynamics Loss and Reward Coefficients to adjustments in Dynamics Randomization or state-observation settings, necessitating retuning under new configurations. We believe this stems from changes in the nonlinearity of the underlying state-to-torque dynamics. We leave a thorough study to future work.

While we have demonstrated our concurrent Intrinsic Dynamics learning and Dynamics Reward mechanism specifically for quadrupedal locomotion tasks, we believe this approach is more generalizable and could produce valuable results in other robotics domains, including bipedal locomotion and manipulation tasks.



\bibliographystyle{IEEEtran}
\bibliography{bibliography}

\end{document}

%% file: combined_torque_prediction_figure.tex
\begin{figure}[t]
    \centering
    \includegraphics[width=\columnwidth]{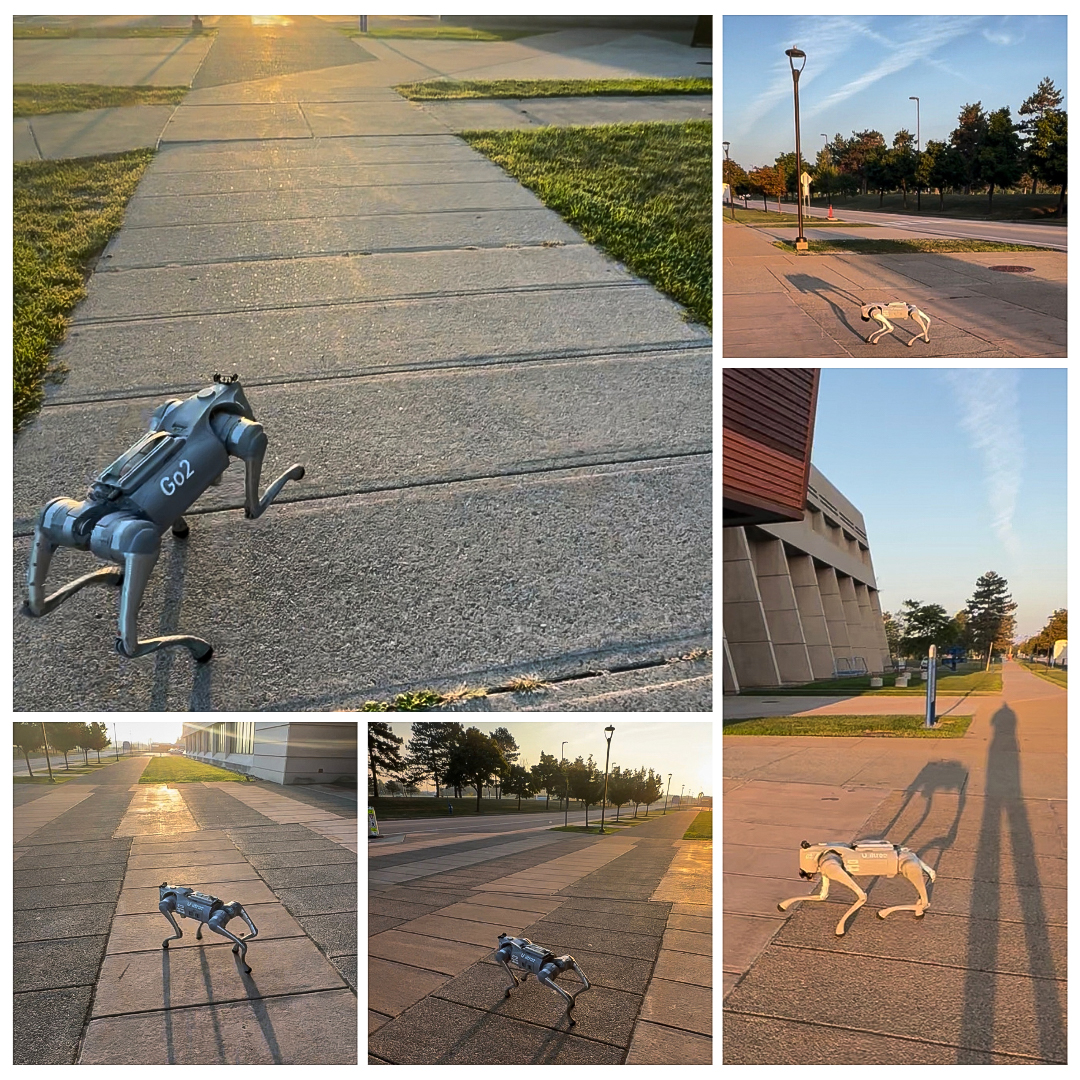}
    \caption{Unitree GO2 Robot navigating over curbs with the Dynamics Aware Controller (videos to be provided subsequently).}
    \label{fig:combined_torque_prediction}
\end{figure}

%% file: privileged_obs_table.tex
\begin{table}[t]
    \centering
    \caption{Privileged Observation Details}
    \label{tab:privileged_obs}
    \begin{tabular}{@{}ll@{}}
        \toprule
        \textbf{Observation Type} & \textbf{Description} \\
        \midrule
        Body Dynamics & Body velocity, COM displacement \\
        \midrule
        Contact Information & Foot contact states, Contact force vectors, \\
        & Contact force magnitudes \\
        \midrule
        Surface Properties & Surface friction, Restitution coefficient, \\
        & Terrain height around feet \\
        \midrule
        Actuator Parameters & Motor strength \\
        \bottomrule
    \end{tabular}
\end{table}

%% file: rewards_active.tex
\begin{table}[t]
\centering
\caption{Active Reward Function Components for Locomotion Policy Training}
\label{tab:reward_new}
\footnotesize
\begin{tabular}{@{}p{2.2cm}p{4.2cm}r@{}}
\toprule
\textbf{Component} & \textbf{Expression} & \textbf{Value} \\
\midrule
\multicolumn{3}{l}{\textit{Velocity Tracking}} \\
\midrule
Linear Velocity & $r_{v} = \begin{cases} \exp(-\frac{(v_{pr} - 0.6)^2}{\sigma_v}) & v_{pr} < 0.6 \\ 1.0 & v_{pr} \geq 0.6 \end{cases}$ & $1.0dt$ \\
\addlinespace[0.1cm]
Angular Velocity & $r_{\omega} = \exp(-\frac{(\omega_{cmd} - \omega_z)^2}{\sigma_{\omega}})$ & $0.5dt$ \\
\midrule
\multicolumn{3}{l}{\textit{Auxiliary Rewards}} \\
\midrule
Orientation & $r_{ori} = \|g_{xy}\|^2$ & $-5.0dt$ \\
\addlinespace[0.1cm]
Z Velocity & $r_{v_z} = v_z^2$ & $-0.02dt$ \\
\addlinespace[0.1cm]
Roll-Pitch Vel. & $r_{\omega_{xy}} = \|\omega_{xy}\|^2$ & $-0.001dt$ \\
\addlinespace[0.1cm]
Base Height & $r_{h} = (\bar{h} - h_{target})^2$ & $-30dt$ \\
\addlinespace[0.1cm]
Collision & $r_{col} = \sum_i \mathbf{1}(\|F_{i}\| > 0.1)$ & $-5.0dt$ \\
\addlinespace[0.1cm]
Foot Slip & $r_{slip} = \sum_i c_i \|v_{f,i}\|^2$ & $-0.04dt$ \\
\addlinespace[0.1cm]
Torque & $r_{\tau} = \|\tau\|^2$ & $-1e-4dt$ \\
\addlinespace[0.1cm]
DOF Pos. Limits & $r_{q_{lim}} = \sum_i \max(0, |q_i| - q_{i,max})$ & $-10.0dt$ \\
\addlinespace[0.1cm]
DOF Velocity & $r_{\dot{q}} = \|\dot{q}\|^2$ & $-1e-4dt$ \\
\addlinespace[0.1cm]
DOF Accel. & $r_{\ddot{q}} = \|(\dot{q}_{t-1} - \dot{q}_t) / \Delta t\|^2$ & $-2.5e-7dt$ \\
\addlinespace[0.1cm]
Power & $r_{P} = \sum_i |\tau_i \dot{q}_i|$ & $-2e-5dt$ \\
\addlinespace[0.1cm]
Action Rate & $r_{a} = \|a_{t-1} - a_t\|^2$ & $-0.01dt$ \\
\addlinespace[0.1cm]
1st Order Smooth. & $r_{s1} = \|q^d_t - q^d_{t-1}\|^2$ & $-0.1dt$ \\
\addlinespace[0.1cm]
2nd Order Smooth. & $r_{s2} = \|q^d_t - 2q^d_{t-1} + q^d_{t-2}\|^2$ & $-0.1dt$ \\
\bottomrule
\end{tabular}
\end{table}

%% file: architecture.tex
\begin{table}[t]
\centering
\caption{Network Architecture for Policy and Adaptation Module}
\label{tab:architecture}
\footnotesize
\begin{tabular}{@{}lcc@{}}
\toprule
\textbf{Component} & \textbf{Architecture} & \textbf{Output Size} \\
\midrule
\multicolumn{3}{l}{\textit{Policy Network}} \\
\midrule
Input & Observation $\mathbf{o}_t$ + Latent $\mathbf{l}_t$ & $n_o + n_l$ \\
Hidden Layer 1 & Linear + ELU & 512 \\
Hidden Layer 2 & Linear + ELU & 256 \\
Hidden Layer 3 & Linear + ELU & 128 \\
Output Layer & Linear & $n_a + n_\tau$ \\
\midrule
\multicolumn{3}{l}{\textit{Adaptation Module}} \\
\midrule
Input & Privileged Information $\hat{\mathbf{s}}_t$ & $n_p$ \\
Hidden Layer 1 & Linear + ELU & 256 \\
Hidden Layer 2 & Linear + ELU & 128 \\
Output Layer & Linear & $n_l$ \\
\midrule
\multicolumn{3}{l}{\textit{Temporal Convolutional Network (TCN) Encoder}} \\
\midrule
Input & Observation History $\mathbf{h}_t$ & $T \times n_o$ \\
TCN Block 1 & 1D Conv($s$=1, $d$=1) + ELU & $T \times 32$ \\
TCN Block 2 & 1D Conv($s$=2, $d$=1) + ELU & $T/2 \times 32$ \\
TCN Block 3 & 1D Conv($s$=1, $d$=2) + ELU & $T/2 \times 32$ \\
TCN Block 4 & 1D Conv($s$=2, $d$=1) + ELU & $T/4 \times 32$ \\
TCN Block 5 & 1D Conv($s$=1, $d$=4) + ELU & $T/4 \times 32$ \\
TCN Block 6 & 1D Conv($s$=2, $d$=1) + ELU & $T/8 \times 32$ \\
Output Layer & Reshape + Linear + LayerNorm & $n_l$ \\
\bottomrule
\end{tabular}
\vspace{0.05cm}
\begin{flushleft}
\footnotesize
$n_o$: observation dimension, $n_l$: latent dimension, $n_a$: action dimension = $n_\tau$: torque dimension, $n_p$: privileged state dimension, $T$: sequence length (100)\\
All TCN blocks use 1D convolutions with kernel size $k$=5 and filter count $f$=32. Parameters: $s$: stride, $d$: dilation rate\\
$n_\tau$ is set to 0 for baseline Policy and is set to 12 (same as $n_a$) for Policy with ID Head
\end{flushleft}
\end{table}

%% file: domain_rand.tex
\begin{table}[!htbp]
\centering
\caption{Domain Randomization Parameters}
\label{tab:domain_rand}
\footnotesize
\begin{tabular}{@{}lc@{}}
\toprule
\textbf{Parameter} & \textbf{Range} \\
\midrule
COM Displacement & $[-0.15, 0.15]$ m \\
Motor Strength & $[0.9, 1.1] \times$ nominal \\
Motor Offset & $[-0.02, 0.02]$ rad \\
Friction & $[0.05, 4.5]$ \\
Restitution & $[0.0, 0.4]$ \\
DOF Position Noise & $0.01$ rad \\
DOF Velocity Noise & $1.5$ rad/s \\
Gravity Noise & $0.05$ m/s$^2$ \\
\bottomrule
\end{tabular}
\end{table}

%% file: dynamics_plots.tex
\begin{figure*}[!htbp]
    \centering
    \begin{subfigure}{\textwidth}
        \centering
        \begin{subfigure}[b]{0.48\textwidth}
            \centering
            \includegraphics[width=\textwidth]{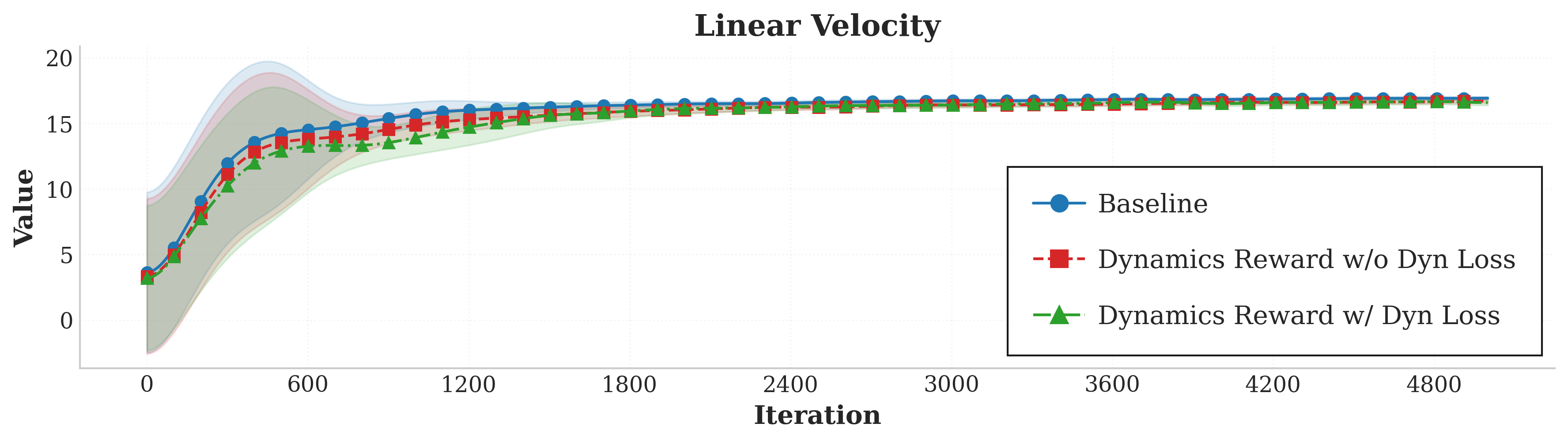}
            \label{fig:lin_vel}
        \end{subfigure}
        \hfill
        \begin{subfigure}[b]{0.48\textwidth}
            \centering
            \includegraphics[width=\textwidth]{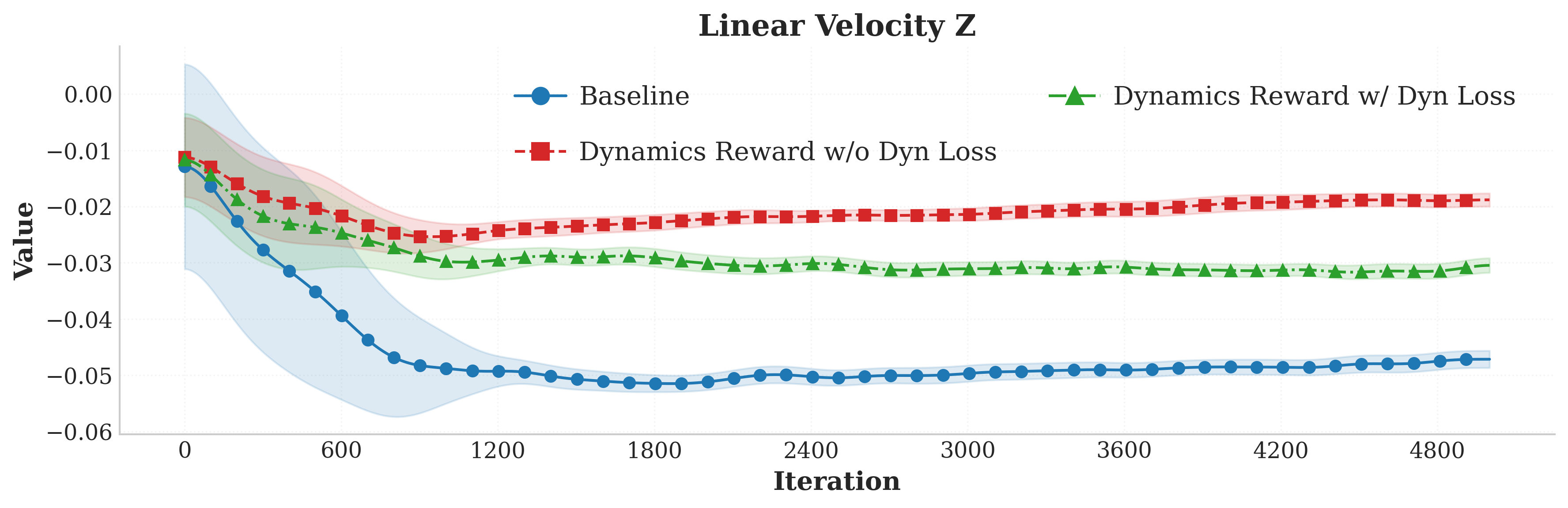}
            \label{fig:lin_vel_z}
        \end{subfigure}
        
        \vspace{0.2cm}
        
        \begin{subfigure}[b]{0.48\textwidth}
            \centering
            \includegraphics[width=\textwidth]{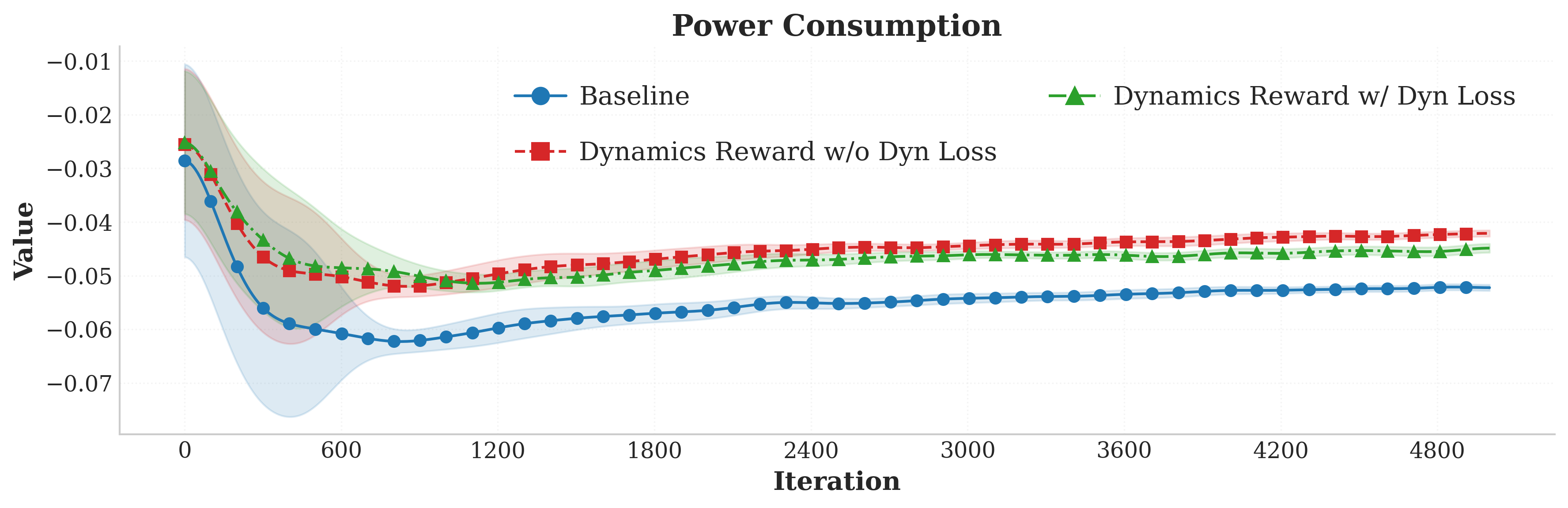}
            \label{fig:power_consumption}
        \end{subfigure}
        \hfill
        \begin{subfigure}[b]{0.48\textwidth}
            \centering
            \includegraphics[width=\textwidth]{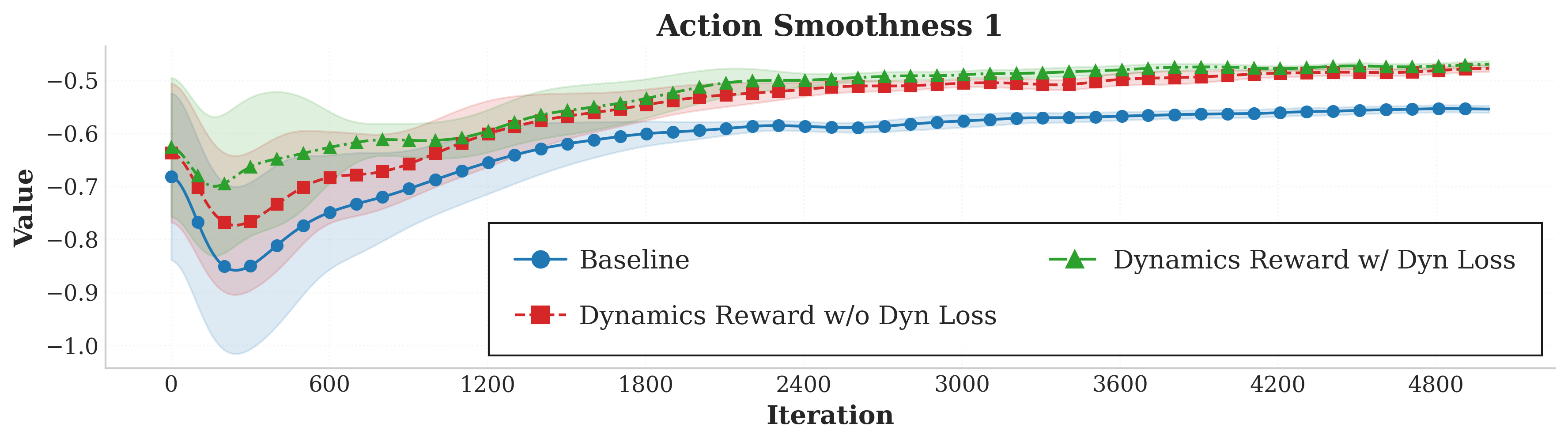}
            \label{fig:action_smoothness}
        \end{subfigure}

        \caption{Performance comparison between baseline controller and proposed approach with Intrinsic Dynamics Head (IDH) and Dynamics Reward, with and without Dynamics Loss. Results averaged over 5 runs. \textbf{Upper Left:} Linear velocity tracking performance remains consistent across all configurations, demonstrating that the proposed approach maintains primary task performance. \textbf{Upper Right and Bottom Row:} Significant improvements in auxiliary metrics (vertical velocity, action smoothness, and power consumption) when using dynamics reward, indicating enhanced controller efficiency, smoothness and stability.}
        \label{fig:subfig_dynamics_reward}
    \end{subfigure}
    
    \caption{Performance analysis of the proposed controller with Dynamics Reward. The plots demonstrate that the proposed approach maintains primary task performance while significantly improving auxiliary metrics related to efficiency and smoothness.}
    \label{fig:combined_results}
\end{figure*}

%% file: prediction_head_rewards_table.tex
\begin{table}[!htbp]
    \centering
    \caption{Reward Improvements with Dynamics Reward}
    \label{tab:prediction_head_rewards}
    \footnotesize
    \begin{tabular}{@{}p{2.8cm}rrr@{}}
        \toprule
        \textbf{Reward} & \textbf{Base} & \textbf{Dynamics} & \textbf{Dynamics} \\
        \textbf{Category} & & \textbf{w/o Aux} & \textbf{w/ Aux} \\
        \midrule
        Tracking Linear Vel & \textbf{16.90} & \underline{16.62} & 16.58 \\
        Tracking Angular Vel & \underline{7.63} & \textbf{7.82} & 7.55 \\
        Linear Velocity Z & -0.047 & \textbf{-0.018} & \underline{-0.030} \\
        Angular Velocity XY & -0.027 & \textbf{-0.023} & \underline{-0.025} \\
        Power Consumption & -0.053 & \textbf{-0.042} & \underline{-0.044} \\
        Action Rate & -1.45 & \underline{-1.27} & \textbf{-1.23} \\
        Action Smoothness 1 & -0.56 & \underline{-0.48} & \textbf{-0.46} \\
        Action Smoothness 2 & -1.09 & \underline{-0.96} & \textbf{-0.90} \\
        DOF Velocity & -0.30 & \underline{-0.25} & \textbf{-0.24} \\
        DOF Acceleration & -1.60 & \underline{-1.34} & \textbf{-1.27} \\
        DOF Pos Limits & \textbf{-0.16} & \underline{-0.20} & -0.21 \\
        Orientation & -0.69 & \textbf{-0.67} & \underline{-0.68} \\
        Base Height & -0.65 & \textbf{-0.54} & \underline{-0.56} \\
        Feet Slip & \underline{-0.16} & \textbf{-0.14} & \underline{-0.16} \\
        Collision & \underline{-0.42} & \textbf{-0.37} & -0.58 \\
        Torques & -0.72 & \textbf{-0.57} & \underline{-0.68} \\
        \bottomrule
    \end{tabular}
    \begin{flushleft}
    \footnotesize
    Results averaged over 5 runs. Higher values are better for tracking rewards, lower (less negative) values are better for penalty rewards. \textbf{Bold} values indicate best performance and \underline{underlined} values indicate second best performance across configurations.
    \end{flushleft}
\end{table}

%% file: real_robot_comparison.tex
\begin{figure}[!t]
\centering

\begin{subfigure}[b]{\columnwidth}
    \includegraphics[width=\columnwidth]{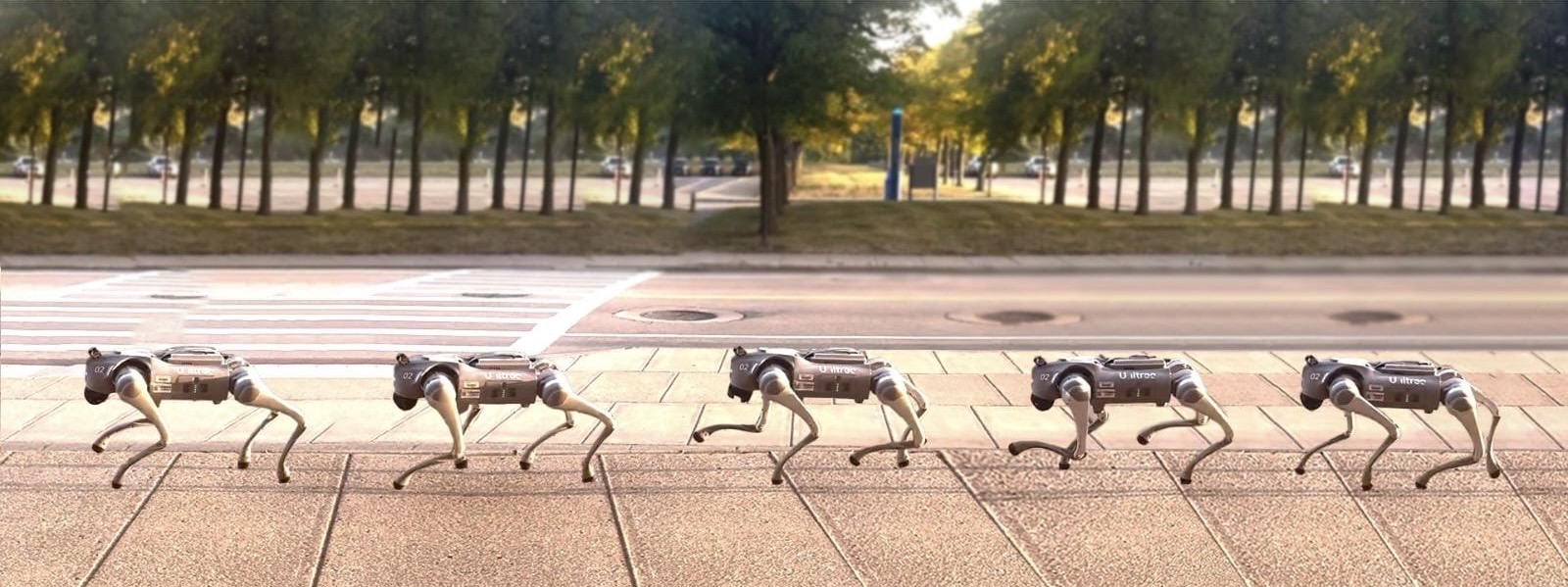}
    \caption{Curb traversal by the Unitree GO2 Robot exhibiting a Trotting gait under the proposed controller.} 
    \label{fig:robot_panorama}
\end{subfigure}

\vspace{0.3cm}

\begin{subfigure}[b]{\columnwidth}
    \includegraphics[width=\columnwidth]{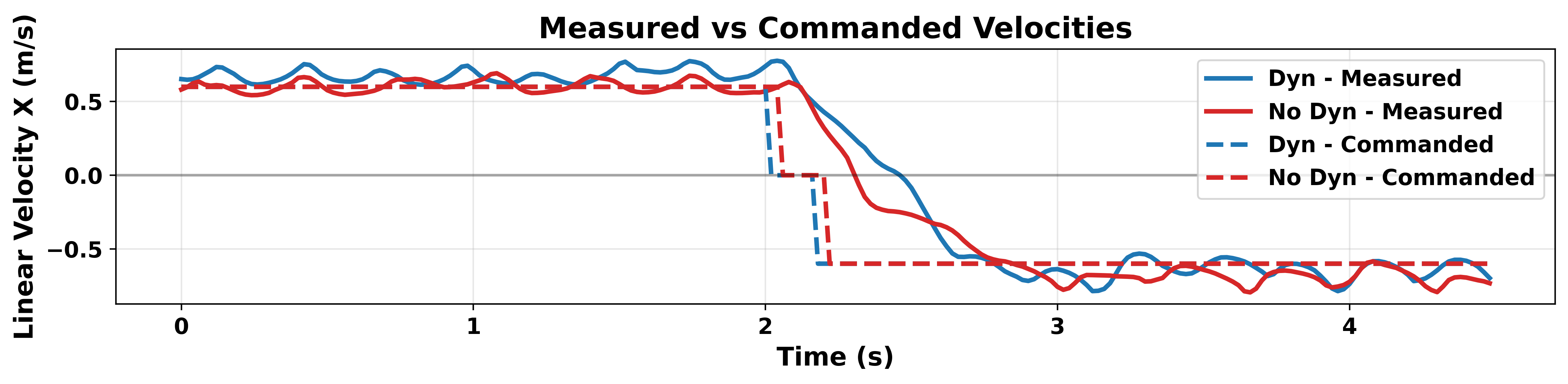}
    \caption{Velocity commands}
    \label{fig:velocity_commands}
\end{subfigure}

\vspace{0.3cm}


\vspace{0.3cm}

\begin{subfigure}[b]{\columnwidth}
    \includegraphics[width=\columnwidth]{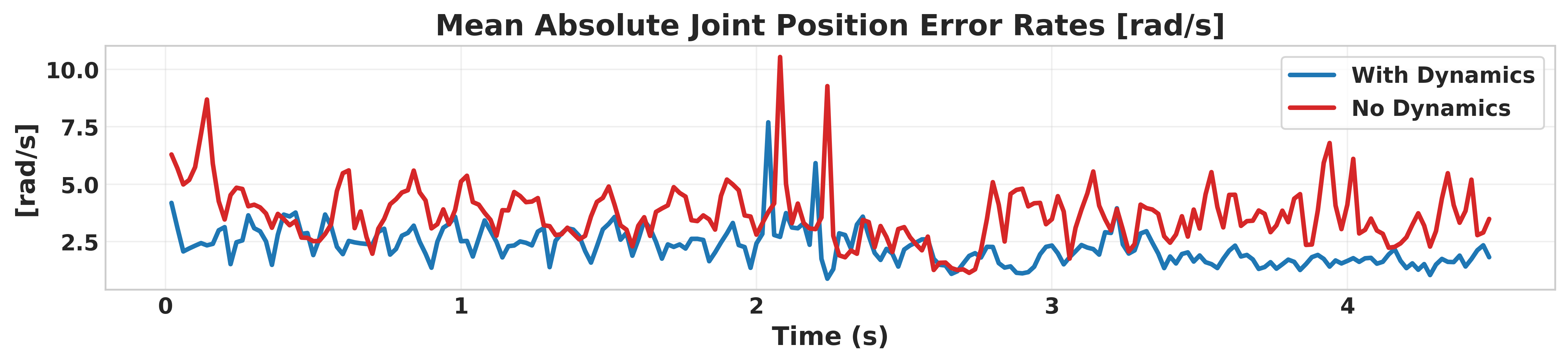}
    \caption{Mean joint position error rate}
    \label{fig:mean_pos_error_rate}
\end{subfigure}

\vspace{0.3cm}

\begin{subfigure}[b]{\columnwidth}
    \includegraphics[width=\columnwidth]{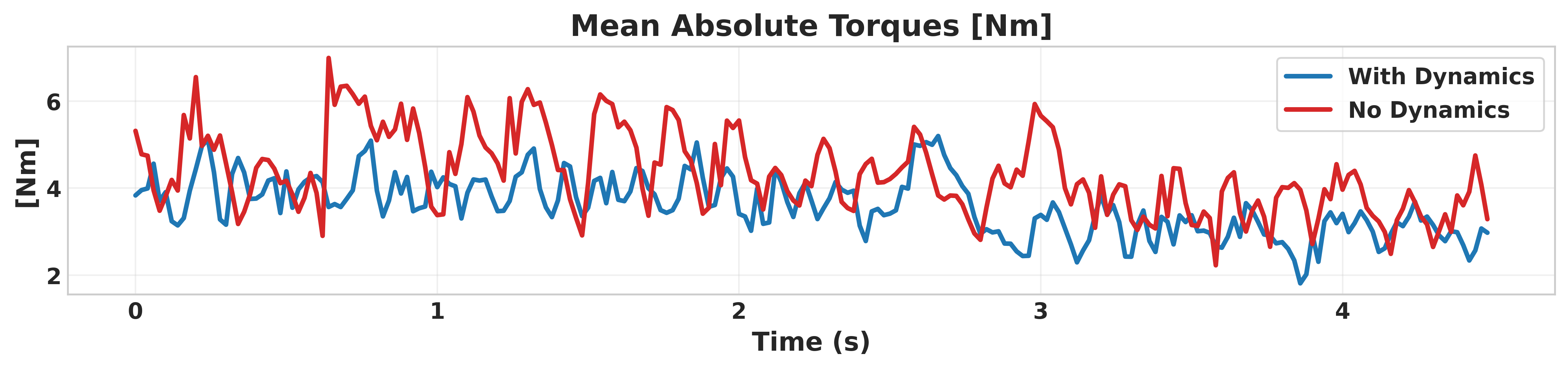}
    \caption{Mean absolute torque}
    \label{fig:mean_abs_torque}
\end{subfigure}

\vspace{0.3cm}

\begin{subfigure}[b]{\columnwidth}
    \includegraphics[width=\columnwidth]{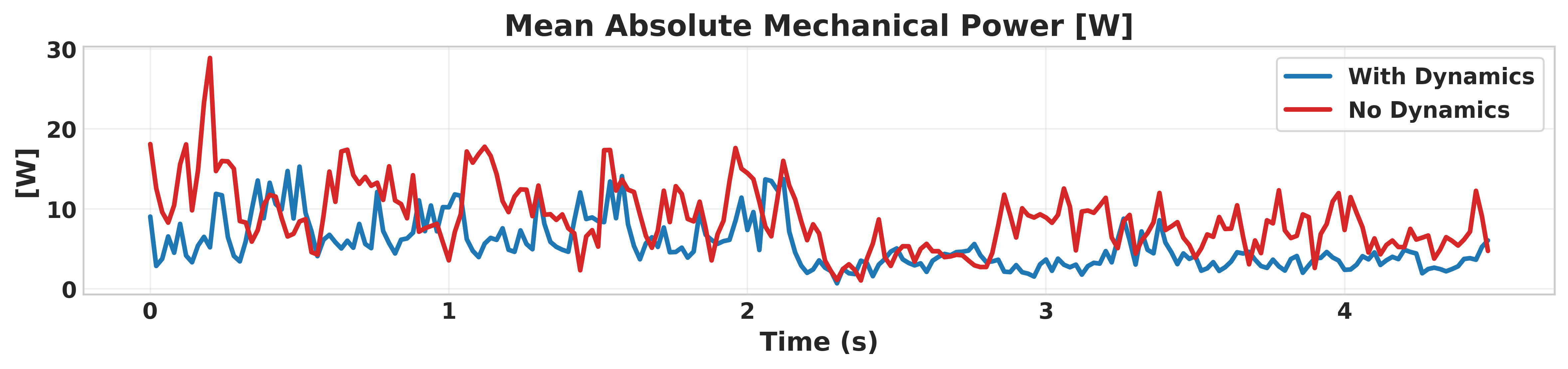}
    \caption{Mean mechanical power}
    \label{fig:mean_power}
\end{subfigure}

\caption{Real robot performance comparison between baseline and dynamics reward policies with identical velocity commands (b). The dynamics reward policy demonstrates improved performance in joint position error rates (c), lower torque requirements (d), and reduced power consumption (e), validating the simulation results in real-world conditions.}
\label{fig:real_robot_comparison}
\end{figure}

%% file: real_robot_metrics.tex
\begin{table}[t]
\centering
\caption{Performance Metrics Comparison on Real Robot. Metrics were computed over 10,000 policy steps of real robot data, averaged across 2 independent runs for each configuration. \textbf{Bold} values indicate best performance, \underline{underlined} values indicate second-best performance, and arrows ($\uparrow$/$\downarrow$) indicate whether higher or lower values are better.}
\label{tab:real_robot_metrics}
\footnotesize
\begin{tabular}{@{}p{3.0cm}@{\hspace{4pt}}p{1.3cm}@{\hspace{4pt}}p{1.3cm}@{\hspace{4pt}}p{1.3cm}@{}}
\toprule
\textbf{Metric} & \textbf{Baseline} & \textbf{Dynamics w/o Aux} & \textbf{Dynamics w/ Aux} \\
\midrule
Position Error RMS [rad] $\downarrow$ & 0.273 & \underline{0.230} & \textbf{0.229} \\
Torque Mean [N·m] $\downarrow$ & 4.591 & \textbf{3.822} & \underline{4.126} \\
Torque Rate RMS [N·m/s] $\downarrow$ & 124.3 & \underline{101.7} & \textbf{96.0} \\
Action Rate RMS [rad/s] $\downarrow$ & 22.6 & \underline{18.4} & \textbf{17.6} \\
Action Acc. RMS [rad/s$^2$] $\downarrow$ & 967.0 & \underline{859.6} & \textbf{684.6} \\
DoF Vel. RMS [rad/s] $\downarrow$ & 3.069 & \underline{2.583} & \textbf{2.550} \\
Mechanical Power [W] $\downarrow$ & 16.680 & \underline{15.450} & \textbf{14.548} \\
Energy Consumption [J] $\downarrow$ & 20681 & \textbf{14596} & \underline{14887} \\
Safe Occupancy Zone \% $\uparrow$ & 88.88\% & \textbf{94.58\%} & \underline{94.31\%} \\
\bottomrule
\end{tabular}
\end{table}